\documentclass{article}

\def\VR#1{{\color{magenta}VR: \it #1}}
\def\VRdel#1{\bgroup\markoverwith{\textcolor{magenta}{\rule[0.5ex]{2pt}{1pt}}}\ULon{#1}}

\usepackage{nips_2017}
\nolinenumbers

\usepackage[utf8]{inputenc} 
\usepackage[T1]{fontenc}    
\usepackage{hyperref}       
\usepackage{url}            
\usepackage{booktabs}       
\usepackage{amsfonts}       
\usepackage{nicefrac}       
\usepackage{microtype}      
\usepackage{multirow}

\usepackage[utf8]{inputenc} 
\usepackage[T1]{fontenc}    

\usepackage{hyperref}       
\usepackage{url}            
\usepackage{booktabs}       
\usepackage{amsfonts}       
\usepackage{nicefrac}       
\usepackage{microtype}      
\usepackage{enumitem}       

\usepackage{amsmath}
\usepackage{amssymb}

\newcommand{\ignore}[1]{}

\newcommand{\squishlist}{
 \begin{list}{$\bullet$}
  { \setlength{\itemsep}{0pt}
     \setlength{\parsep}{3pt}
     \setlength{\topsep}{3pt}
     \setlength{\partopsep}{0pt}
     \setlength{\leftmargin}{1.5em}
     \setlength{\labelwidth}{1em}
     \setlength{\labelsep}{0.5em} } }
\newcommand{\squishend}{
  \end{list}  }

\usepackage{graphicx} 
\usepackage{xcolor}
\usepackage[normalem]{ulem}

\def\comment#1{{\color{gray} \it #1}}

\def\VR#1{{\color{magenta}VR: \it #1}}
\def\VRdel#1{\bgroup\markoverwith{\textcolor{magenta}{\rule[0.5ex]{2pt}{1pt}}}\ULon{#1}}

\def\OLdel#1{\bgroup\markoverwith{\textcolor{blue}{\rule[0.5ex]{2pt}{1pt}}}\ULon{#1}}

\def\IPdel#1{\bgroup\markoverwith{\textcolor{brown}{\rule[0.5ex]{2pt}{1pt}}}\ULon{#1}}

\definecolor{darkgreen}{rgb}{0.0, 0.5, 0.0}

\def\ODdel#1{\bgroup\markoverwith{\textcolor{darkgreen}{\rule[0.5ex]{2pt}{1pt}}}\ULon{#1}}

\definecolor{purple}{rgb}{0.5, 0.0, 0.5}

\def\ACdel#1{\bgroup\markoverwith{\textcolor{purple}{\rule[0.5ex]{2pt}{1pt}}}\ULon{#1}}

\definecolor{darkred}{rgb}{0.5, 0.0, 0.0}

\def\JPdel#1{\bgroup\markoverwith{\textcolor{darkred}{\rule[0.5ex]{2pt}{1pt}}}\ULon{#1}}

\title{Social Dialogue using an Ensemble Model with Neural Ranking: What Users Like about Alana}
\title{What Users Like about Social Dialogue Systems}
\title{An Ensemble Model
with Ranking for Social Dialogue}

\author{
	 Ioannis Papaioannou, Amanda Cercas Curry, Jose L. Part\\
     \textbf{Igor Shalyminov, Xinnuo Xu, Yanchao Yu}\\
     \textbf{Ondřej Dušek, Verena Rieser, Oliver Lemon}\\ 
	The Interaction Lab,
    Department of Computer Science\\
	Heriot-Watt University, 
	Edinburgh, EH14 4AS, UK\\
    \texttt{\{ac293, v.t.rieser\}@hw.ac.uk} 
    }

\begin{document}

\maketitle

\begin{abstract}
Open-domain social dialogue is one of the long-standing goals of Artificial Intelligence.
This year, the Amazon Alexa Prize challenge was announced for the  first time, where real customers get to rate systems developed by leading universities worldwide. The aim of the challenge is to converse ``coherently and engagingly with humans on popular topics for 20 minutes''.
 
 We describe our Alexa Prize system (called `Alana') consisting of an ensemble of bots, combining rule-based and machine learning systems, and using a  contextual ranking mechanism to choose a system response.
 The ranker was trained on real user feedback received during the competition, 
 where we address the problem of
 how to train on the noisy and sparse feedback obtained during the competition.

 
\end{abstract}

\section{Introduction}\label{sec:intro}

This paper discusses two of the major challenges when building open-domain social dialogue systems:
\begin{enumerate}
\item How can we facilitate  open domain interaction while still executing control?
\item Which utterance fits best in a given dialogue context?
\end{enumerate}

Early systems for social chat, such as ELIZA \citep{eliza},  were based on carefully handwritten rules, but recent systems  are often trained using a variety of (deep) learning techniques over large public data sets, such as OpenSubtitles or Twitter \citep[e.g.][]{vinyals2015neural,sordoni-EtAl:2015:NAACL-HLT,Li2016}.
However, learning directly from data also has its pitfalls when deploying a system to real customers, as recent examples such as Microsoft's Tay bot
demonstrate.
We present a hybrid model, incorporating hand-crafted rules (validated and developed through customer feedback) and machine learning models trained on carefully chosen datasets.

Following previous hybrid systems, \cite[e.g.][]{yu_strategy_2016}, we apply a ranker model to select the most relevant reply from a pool of replies generated by an ensemble of different agents/bots. 
It is still an open question how to best define this ranking function. Previous work has manually defined a evaluation function 
based on hand-selected turn-level features \citep{yu_strategy_2016,Li2016}. 
Other work has experimented with learning from crowdsourced user ratings \citep{lowe:acl2017}.
One major drawback of such previous work is that it only evaluates a possible response locally, i.e.\ per turn, rather than considering its contribution to the overall dialogue outcome, (e.g.\ to engage the user. As such, these ranking functions often favour safe, but dull responses \citep{lowe:acl2017}). 
We experimented with a variety of ranking functions and datasets as described below. 
This resulted in one of the top bots in the competition according to average customer rating, as well as with respect to average dialogue length.

\section{System Design and Architecture}\label{sec:system}

\begin{figure}[!htb]
	\centering
		\includegraphics[width=\linewidth]{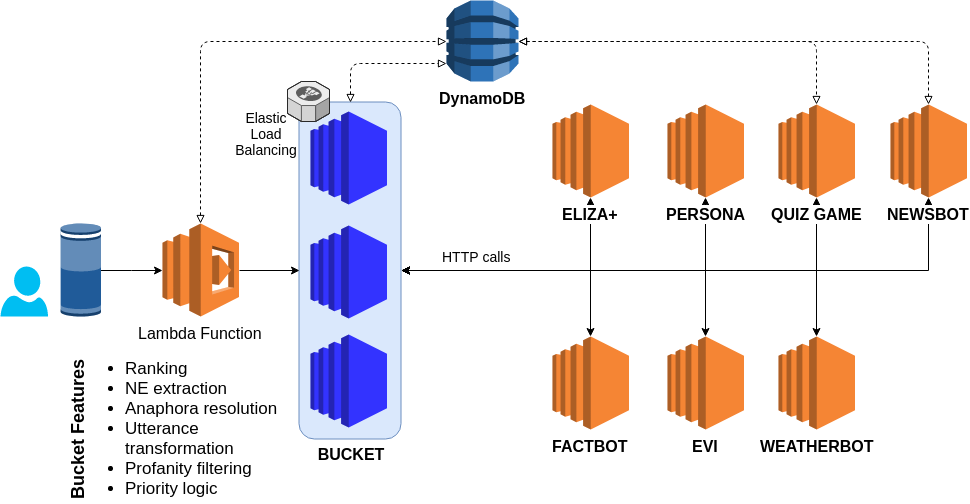}
		\rule{35em}{0.5pt}
        \caption[Alana/ Watt's up Architecture]{Alana is a hybrid hierarchical architecture with ranking}  \label{fig:alexa_architecture}
\end{figure}

The system architecture is shown in Fig.~\ref{fig:alexa_architecture}.
We rely on an ensemble of bots. These bots fall into two main categories:

\begin{enumerate}[itemsep=0pt,topsep=0pt,leftmargin=10pt,labelwidth=10pt,label=\textbf{\arabic*.}]
\item {\bf Data-driven Bots:} We experimented with retrieval based bots as well as generative Sequence-to-Sequence models (Seq2Seq, see section \ref{sec:seq2seq}) While the former always produce well-formed sentences (as retrieved from the data set), the latter can generate new and possibly more contextually appropriate replies, however at the expense on needing larger data sets to learn from. We follow previous work by combing both paradigms into an ensemble-based approach \citep{SongYLZZ:2016}.
\item {\bf Rule-based bots} are used to respond to the specific user queries in a controlled and consistent way, (e.g.\ to queries about the personality of our bot, such as favourite things etc., or the weather), using a combination of in-house  developed bots and extended versions of 3rd party bots. 
\end{enumerate}

These two categories include the following bots:

{\bf Persona:} A rule-based system implemented in AIML\footnote{\url{http://www.alicebot.org/aiml.html}} whose main purpose is to maintain personality-based responses consistent across turns, such as music tastes or other preferences. \textit{Persona} also includes replies to other topics, where we want to guarantee an appropriate response to inappropriate user utterances and topics such as sex, as per the competition rules.

{\bf Eliza:} We extended an existing 
Eliza-style chatbot called \textit{Rosie}.\footnote{\url{https://github.com/pandorabots/rosie}} Since the initial \textit{Rosie} bot was designed for 
 mobile devices, we heavily altered it for the Challenge.

{\bf NewsBot:} 
An information retrieval bot based on an open-source framework Lucene.\footnote{\url{https://lucene.apache.org}} We build and continuously populate a search index 
of selected news sources provided via NewsAPI.\footnote{\url{https://newsapi.org}} 
For indexing as well as for the bot's responses, we use summaries of the news articles extracted with an open-source library called {\em Sumy}.\footnote{\url{https://pypi.python.org/pypi/sumy}} 
In order to select a relevant piece of news for a user's query, 
we create 1, 2, and 3-grams over the user's utterance and dialogue context.
We employ the BM25 algorithm to score news relevance,
with named entities and noun phrases from the user query boosted using a set of weights adjusted empirically. A re-ranking step is then applied for the top 10 candidates based on the articles' recency.

{\bf Factbot -- Fun facts, Jokes, and Stories:} A collection of facts, jokes and stories that get triggered whenever the user  specifically asks for them or as a deflection strategy when no suitable response is found. 
For the fun facts, the user can also specify a named entity (``{\em Tell me a fact about X}''). 
 Otherwise, a 
 fact is chosen randomly. The data was collected from a multitude of online resources.

{\bf Quiz Game:} A hand-crafted system developed using a VoiceXML-based structure. During the game, the user has to guess the right answer to topic-specific questions 
 (e.g.\ 80's music, science, history, sport and geography). 
The user can end the game at any point. 

{\bf EVI:} A third party bot retrieving factual information (if applicable) about the user utterance, powered by the EVI  question answering engine API.\footnote{\url{https://www.evi.com/})} This bot returns only one candidate if there is one. Some  EVI answers which would not be appropriate in a dialogue are filtered out.

{\bf Weatherbot:} A simple rule-based bot that provides the user with weather-related information, if asked for, querying the \textit{OpenWeatherMap API}\footnote{\url{https://openweathermap.org/}} on the fly.

Each of these bots produces a possible system utterance according to its internal rules. Note that not all bots fire at each turn.
All the returned candidates are postprocessed and normalized. 
Profanity, single-word and repetitive (news only) candidates are filtered out. 
The final system response is selected in three steps:
\begin{enumerate}[itemsep=0pt,topsep=0pt,leftmargin=10pt,labelwidth=10pt,label=\textbf{\arabic*.}]
\item \textbf{Bot priority list.} Some of the deployed bots are prioritized, i.e.\ if they produce a response, it is always selected. The priority order is the following: \textit{Quiz game, Factbot, Weatherbot, Persona, Evi}. 
\item \textbf{Contextual priority.} The NewsBot's response is prioritized if it stays on the topic of a previously mentioned news story.

\item \textbf{Ranking function.} If none of the priority bots produced an answer, the rest of the deployed bots' responses populate the list of candidates and the best response is selected via a ranking function, see Section \ref{sec:devel}.
\end{enumerate}
In the extreme case where none of the bots produced an answer (or all of them were filtered out due to postprocessing rules), the system returns a random fun fact, produced by the \textit{Factbot}.
Please refer to \cite{alexa_2017} for more details.

\subsection{Other Bots and Data}

We also experimented with other data-driven bots, which were not included in the final system.

\subsubsection{Data Sets for Information Retrieval Bots}\label{ssec:data}




\begin{itemize}[itemsep=0pt,topsep=0pt,leftmargin=10pt,labelwidth=10pt]
\item\textbf{OpenSubtitles} \citep{lison_opensubtitles2016:_2016}, with the automatic turn segmentation provided by \citet{lison_automatic_2016}. We used all dialogues of two or more turns and filtered the data as described below. 

\item\textbf{Cornell Movies, Jabberwacky, CNN:} these datasets proved to be too small for our purposes: Cornell Movie Dataset \citep{danescu-niculescu-mizil_cornell-movies_2011}, Jabberwacky chatbot chat logs\footnote{\url{http://www.jabberwacky.com/}}, and CNN chat show transcripts from \cite{yu_strategy_2016,yu_learning_2017}.
\end{itemize}

In order to comply with the competition rules, we first filtered the data for profanities.
However, profanities are often context-dependent 
 and hard to capture by a purely lexicon-driven approach. 
As such, we experimented with restricting the OpenSubtitles data set using age ratings of the movies.
We obtained movie ratings from IMDb
and only included in our dataset the movies with a U.S.\ “G” or U.K.\ “U” ratings (“general”, “universal”).

Another problem from OpenSubtitles data was the occurrence of many personal names and other named entities that would appear out-of-context in a dialogue. We used Stanford NER \citep{finkel_stanford-ner_2005} to detect named entities and filtered out all context-response pairs containing named entities in the response.
However, the downside of this approach is that we ended up with much smaller data sets 
which made data-driven approaches, such as  the  generative Seq2Seq approach less feasible.

\subsubsection{Seq2Seq}\label{sec:seq2seq}
Throughout    system development, we experimented with a sequence-to-sequence dialogue model \citep{vinyals2015neural}, training it on several datasets. 
The first promising behaviour was obtained with Twitter data\footnote{\url{https://github.com/Marsan-Ma/chat_corpus}}: it was interesting and mostly grammatical yet often offensive and politically related.  We then switched to a subset of Reddit logs~-- over 21,000 conversation snippets in the form of question-answer pairs cleaned from profanity and filtered to only contain small-talk conversation (thanks to Dr.\ Zhuoran Wang).
In order to exclude ungrammatical responses, we disregarded all  answers with a low confidence score (defined as the sum of the logits at the decoder's output). We adjusted the confidence threshold empirically on a separate development set of 100 sample user utterances both collected from WoChat\footnote{\url{http://workshop.colips.org/wochat/}} transcripts and paraphrased from a list of popular daily topics provided by Amazon.

The experiment thus resulted in a casual conversation bot: its answers are supposed to be given at times when the user is following up on the previous system's answer or just hesitating. 
Due to  time constraints, the final version of the seq2seq bot was not deployed into production, and so its possible contribution to the users' ratings is left for future work.

\section{Example Dialogue}\label{sec:examples}
Note: The dialogue presented here  does not come from real customer data, but was recreated  by interacting with our system (running a text-based version on Telegram) The same structure of interaction and   Named Entities are used as occurred in a real dialogue.  We also indicate which bot in our ensemble generated the answer.

\ignore{ 
\noindent\fbox{%
    \parbox{\textwidth}{%
U: can we talk about music\\
A:  \comment{(Persona)} Who is your favorite band?\\
U: the Doors\\
A:  \comment{(Evi)} The Doors were an American rock band formed in 1965
in Los Angeles, with vocalist Jim Morrison, keyboardist Ray Manzarek,
guitarist Robby Krieger, and drummer John Densmore.  What you think
about Los Angeles.\\
U: it's an okay place\\
A : \comment{(Eliza)} I can see you've put thought into that.\\
U: and what are your views on it\\
A: \comment{(News API)} I saw this on the news, Los Angeles
strikes deal to host 2028 Summer Olympics: Source. Do you want to
learn more?\\
U: yes sure\\
A: \comment{(Multi-turn News)} I'm so happy that you are interested in this topic.
Here's more. Then in June, the International Olympic Committee decided
that with two strong candidates, Los Angeles and Paris, it would give
one city the 2024 games and the other the 2028 games. This will be Los
Angeles' third time hosting the Olympics. What do you think about Los
Angeles? \VR{Repetition!}\\ 
U: ...
    }%
}
 

\noindent\fbox{%
    \parbox{\textwidth}{%
U:  so  have you ever played a game\\
A: \comment{(Eliza?)} No I didn't. What's like?\\
U:  I can tell  you about my favorite game Minecraft\\
A: \comment{(News API)}  Did you see the news about Minecraft is
now available for cross-play on any device. With the launch of
Microsoft's new Xbox One X gamers got an update on the latest
Minecraft news -- chiefly that Minecraft is now available across any
device. Do you want to know more?\\
U: yes thanks  for telling me that because I didn't know that\\
A:  \comment{(Multi-turn News)} That's all right. Let's talk about it more. The new
update also features better graphics and cross-platform support to let
Windows 10 and mobile Minecraft players play the game together. So a
pack bought on mobile will also be available on Xbox and Windows 10
editions too. I was wondering what you think about Minecraft.\\
U: I love it it's a really fun game ....\\
A: ...
 }%
}

\noindent\fbox{%
    \parbox{\textwidth}{%
    
U: is there any gossip\\
A :  Do you think Miley Cyrus and Liam Hemsworth will last?\\
U:  how old is she \comment{(Anaphora resolution)}\\
A  Miley Cyrus is 24 years old.\\
User:  well is there any news about her \comment{(Anaphora resolution)}\\
 A: I've been reading the news and I saw an article about Miley
Cyrus on toning down her image: 'I did feel sexualized'. Miley Cyrus
has said goodbye to her tongue-wagging, twerking days. Do you want to
learn more?\\
U: sure\\
A: I'm so happy that you are interested in this subject. Here's
more. The Malibu singer said she decided to tone down her image after
her sometimes shocking antics left her feeling exploited. It became
something that was expected of me, she said of her over-the-top
fashion sense and wild persona in a new cover story for Harper's
Bazaar. What are your views on Miley Cyrus?\\

 }%
}
} 

\noindent\fbox{%
    \parbox{\textwidth}{%
  
A: Hi, [...]
What would you like to talk about? \\
U: music \\
A: \comment{(Persona)} Great! Who is your favorite singer? \\
U: Bob Dylan \\
A: \comment{(Evi + additional question)}  Bob Dylan is an American songwriter, singer, painter, and writer. What are your opinions on Bob Dylan?\\
U: When was {\bf he} born? \\ 
A: \comment{(Evi + anaphora resolution)} Bob Dylan was born on Saturday May 24th 1941. \\
U: What's happening with {\bf him}?\\
A: \comment{(News-multi-turn + anaphora resolution)} I heard this on the radio -- Was Dylan too freewheelin' in borrowing for Nobel lecture?  The whiff of plagiarism is blowin' in the wind for Bob Dylan. Want to know more?\\
U: sure\\
A: \comment{(News-multi-turn)}  It seems that you are interested in this topic. I'm so glad. Here's more. Here's Dylan: Some men who receive injuries are led to God, others are led to bitterness The cases Pitzer found are not blatant or explicit -- there are no verbatim sentences, only identical phrases and similar phrasing. What are your opinions on Bob Dylan?\\
U: I love him! \\
A: \comment{(Persona)} Great! What else do you like?  \\ 
U: cooking \\
A: \comment{(Facts)} I heard that every day, Americans eat an estimated 18 acres of pizza.
}}

\section{Experiments with Ranking Functions}\label{sec:devel}

The responses proposed by each   bot are ranked according to a set of features. 
We have experimented with several ranking functions. 

\subsection{Hand-engineered Ranker function}\label{sec:handcrafted-ranker}

The hand-engineered ranking function    uses the following features:
\begin{itemize}[itemsep=0pt,topsep=0pt,leftmargin=10pt,labelwidth=10pt]
\item \textbf{Coherence:} Following \cite{Li2016}, we reward semantic similarity between the user's utterance and the candidates using Word2Vec \citep{word2vec} 
\item \textbf{Flow:} Also similar to \cite{Li2016}, we penalise similarity between consecutive system utterances in order to prevent repetition. Here, we use both Word2Vec 
and METEOR word n-gram overlap 
as measures of similarity.
\item \textbf{Questions:} 
By promoting  questions, we aim to incite the user to continue the conversation. 
\item \textbf{Named Entities:} We strongly reward utterances containing the same named entities as the user's reply to promote candidates relating to the same topic. 
\item \textbf{Noun Phrases:} Similarly, we reward matching noun phrases between the user's and the system's utterances. Noun phrases are identified based on part-of-speech tagging.
\item \textbf{Dullness:} We compare each response to a list of dull responses such as ``I don't know" and penalise Word2Vec similarity between them, since we would like the bot's utterances to be engaging, similarly to \cite{Li2016}.
\item\textbf{Topic Divergence:} We trained a Latent Dirichlet Allocation (LDA) model on a weighted combination of preprocessed versions of the OpenSubtitles and the WashingtonPost datasets. We set the vocabulary size to $20k$ and the number of topics to $200$, and we used a tailored stop-words list. For every proposed answer in the bucket, we compute the topic divergence from the user utterance. 

\item\textbf{Sentiment Polarity:} We use the VADER sentiment analyser \citep{gilbert_vader:_2014} from the NLTK toolkit,\footnote{\url{http://www.nltk.org/api/nltk.sentiment.html}} which provides a floating point value indicating sentence sentiment.
\end{itemize}

These features are calculated using the last two system turns in order 
 to maintain  dialogue context.
The final score is a weighted sum of these features:

\begin{equation}
\begin{split}
	score = 0.25*turn_0 + 0.25*turn_1 + 0.25*turn_2 + 0.25*noun\_phrases\\
	+ 3*named\_entities - 0.25*topic\_divergence
\end{split}
\end{equation}

where  $turn_i$ is computed 
using the $i$-th utterance counting from the end of the dialogue history:

\begin{equation}
\begin{split}
	turn_i = -0.2*flow_{sem\_similarity} - 3*flow_{METEOR} + 0.1*coherence_{sem\_similarity}\\
    - 0.24*dullness + 0.2*question + 0.1*sentiment\_polarity
\end{split}
\end{equation}






\subsection{Linear Classifier Ranker}\label{sec:linear-ranker}

In order to use the feedback ratings obtained from real users in the competition, we also trained the VowpalWabbit linear classifier \citep{langford_vowpal_2007} to rank Bucket responses based on the following features:
\begin{itemize}[itemsep=0pt,topsep=0pt,leftmargin=10pt,labelwidth=10pt]
\item bag-of-n-grams from the context (preceding 3 utterances) and the response (unigrams, bigrams, and trigrams)
\item position-specific n-grams at the beginning of the context and the response (first 5 positions)
\item dialogue flow features, same as for the hand-engineered ranker (see Section~\ref{sec:handcrafted-ranker})
\item bot name.
\end{itemize}
The ranker is trained as a binary classifier, but it outputs a floating-point score in practice. At runtime, 
the highest-scoring response is selected for the output.

We initially trained the ranker on Cornell movies, Twitter, and Jabberwacky datasets (see Section~\ref{ssec:data}), with positive examples from the real dialogues and negative ones randomly sampled from the rest of the set, but the ranker only learned to prefer responses similar to data from these datasets; its performance in real dialogues was lacking in our tests.
Therefore, after collecting enough live dialogues during the Alexa Prize competition, we retrained the ranker on {\it real dialogues collected during the competition}. The rating target function is an approximation of human ratings – we use all context-response pairs from successful dialogues (human rating 4 or 5) as positive examples (value +1) and all pairs from unsuccessful dialogues (rating 1 or 2) as negative (value -1) and train the ranker to mimic this rating. 

We collected 60k dialogue instances over one month for training and 7k dialogue instances over 4 days as a development set.
We did not perform any large-scale parameter optimization, but based on performance on the development data, we selected the following VowpalWabbit parameters: \begin{itemize}[itemsep=0pt,topsep=0pt,leftmargin=10pt,labelwidth=10pt]
\item logistic loss function (logistic regression),
\item feature concatenations (context + response n-grams, pairs of n-grams from responses, bot name + response n-grams, bot name + context n-grams, bot name + dialogue flow, bot name + context n-grams + response n-grams), 
\item 16-bit feature hash table,
\item 1 pass over the training data. 
\end{itemize}
This setup reached 69.40\% accuracy in classifying the development data items as positive or negative.
The results of deploying this Linear Ranker are presented in section \ref{results:linear}.

\subsection{Results}\label{results:linear}

The Linear Ranker, trained on the user feedback received during the competition (see Section \ref{sec:linear-ranker}), was deployed  on top of Alana v1.1, and evaluated in comparison to the    hand-crafted ranking function (see Section~\ref{sec:handcrafted-ranker}).
The results are shown in Table~\ref{tab:results-linear}.
\begin{table*}[ht]
\begin{center}
\begin{tabular}{|c||c|c|}\hline
{\bf System} & average user rating  & number of dialogues  \\\hline \hline
  
Alana v1.1  :  Hand-engineered Ranker&3.26 & 191  \\\hline 
Alana v1.1 : Trained Linear Ranker &3.28& 272  \\\hline 
\end{tabular}
\end{center} \caption{Results: Trained Linear Ranker (semi-finals period)}\label{tab:results-linear}
\end{table*}

This shows that we can continuously improve system performance by training on real customer feedback from the competition, even though it is noisy and sparse (ratings are only available for whole dialogues, and not each dialogue turn).

\section{Future Work}
This paper describes our Alexa system as entered in the semi-finals (July-August 2017).
We are now competing as one of three systems  in the Amazon Alexa Challenge finals, where we have replaced the linear ranker with a neural model. This neural ranker is trained on an increased number of user ratings, which we were able to gather August-October 2017, and outperforms the linear ranker in terms of accuracy.

\subsubsection*{Acknowledgements}
We would like to thank Helen Hastie and Arash Eshghi for their helpful comments and discussions.

\small

\bibliographystyle{apalike}
\bibliography{alexa}

\end{document}